\documentclass[runningheads]{llncs}

\usepackage[T1]{fontenc}
\usepackage{graphicx}
\usepackage{amsmath,amssymb}
\usepackage{booktabs}
\usepackage[hidelinks]{hyperref}

\begin{document}
\title{VDSB-GWSyn: Diffusion Schr\"{o}dinger Bridge for Controllable and Anatomically Feasible Guidewire Synthesis in Coronary Angiography}
\titlerunning{VDSB-GWSyn for Guidewire Synthesis}
%
\author{Haoyuan Tang\inst{1}\orcidID{0009-0002-3107-569X} \and
Zhuo Zhang\inst{1}\orcidID{0000-0002-3946-0720}\thanks{Corresponding author.} \and
Jialin Li\inst{1}\orcidID{0009-0007-2608-1565} \and
Shuai Xiao\inst{1}\orcidID{0000-0003-4058-8120} \and
Jiachen Yang\inst{1}\orcidID{0000-0003-2558-552X}}


\authorrunning{Tang et al.}

\institute{Tianjin University, Tianjin, China\\
\email{z\_zhuo@tju.edu.cn}}

\maketitle              
\begin{abstract}
Coronary guidewire endpoint localization is a fundamental capability for computer-assisted PCI, and its importance increases as robot-assisted PCI is progressively adopted to reduce operator radiation exposure. However, the scarcity of annotated CAG images with guidewires and the limited adaptability of existing guidewire synthesis models remain key bottlenecks for guidewire endpoint localization. To address this issue, we propose VDSB-GWSyn, a Diffusion Schr\"{o}dinger Bridge (DSB) model-based framework, enabling synthesis of controllable, high-fidelity guidewire samples under complex anatomical backgrounds. VDSB-GWSyn first uses our shape prior algorithm to learn the basic guidewire geometry. It then generates guidewire masks under constraints imposed by the vessel segmentation masks and outputs the corresponding endpoint coordinates. Finally, it synthesizes realistic guidewire samples on real CAG images using DSB conditioned with SPADE. Experimental results show that the guidewire samples synthesized by VDSB-GWSyn achieve favorable ROI-FID and ROI-KID, as well as high IPR scores. In addition, incorporating our synthesized data for synthetic pre-training followed by real fine-tuning substantially improves downstream guidewire endpoint localization, reducing MPE from 16.01~px to 7.71~px and increasing PCK at 3~px from 52.63\% to 86.27\%, leading to more clinically reliable deployment of robot-assisted guidewire delivery systems. Moreover, the core design philosophy of controllable device synthesis with strict background preservation and anatomical feasibility constraints has the potential to transfer to other interventional device perception tasks where annotated data are scarce.

\end{abstract}
\begin{figure}[t]
\centering
\includegraphics[width=\textwidth]{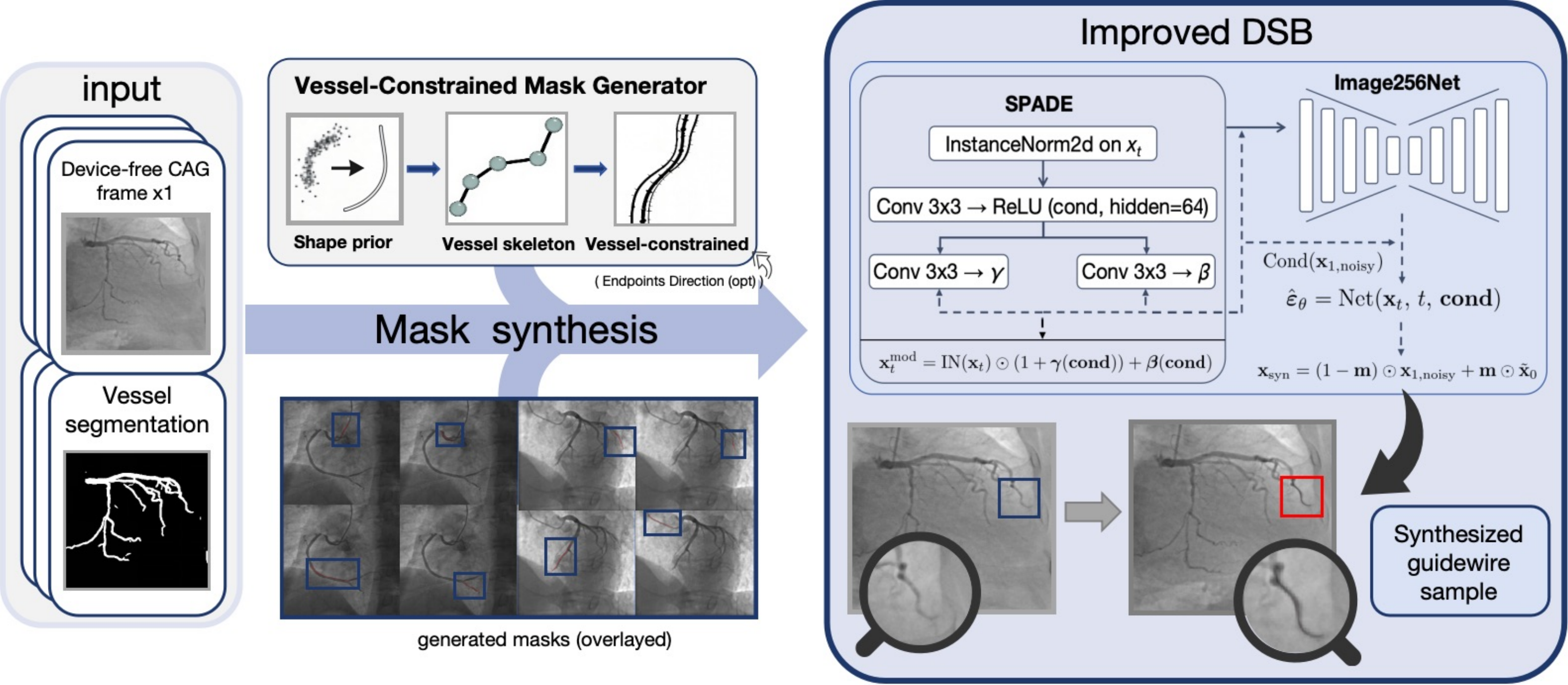}
\caption{Overview of the proposed VDSB-GWSyn framework for controllable guidewire synthesis.}
\label{fig:pipeline}
\end{figure}
    
\section{Introduction}
Reliable guidewire perception is a prerequisite for computer aided navigation in coronary angiography~\cite{li2021unified}. In fluoroscopy, the distal tip is often weakly visible and easily confounded by complex anatomical backgrounds~\cite{pan2025label}, making endpoint localization difficult and error prone~\cite{li2021real}. Deep learning based endpoint localization requires large scale, high quality annotations, yet labeling guidewire pixels and tip coordinates is labor intensive and privacy constraints further limit data collection and sharing~\cite{dorjsembe2024conditional,yazdani2025flow}. As a result, data scarcity remains a central bottleneck for robust endpoint localization in real world fluoroscopy. 
    
Prior work has explored learning-based catheter and guidewire segmentation under limited annotations and augmentation, and has also investigated synthetic fluoroscopic data and transfer learning to reduce labeling requirements~\cite{ambrosini2017fully,gherardini2020catheter}. For image synthesis, unpaired adversarial translation has been applied to catheter synthesis in X-ray fluoroscopy to reduce pixel-level labeling effort, and adversarial augmentation has been used to improve catheter segmentation~\cite{ullah2021synthesize,ullah2019catheter}. More recently, diffusion-based generators have demonstrated improved controllability through mask conditioning, and have been used to synthesize labeled fluoroscopy videos for guidewire segmentation~\cite{dorjsembe2024conditional,du2023arsdm,konz2024anatomically,pan2025label}. Despite these advances, two challenges remain for guidewire endpoint localization. First, existing conditional synthesis pipelines typically assume that a valid guidewire mask is already available as input, while the anatomical feasibility of the guidewire trajectory within coronary vessels is not explicitly enforced inside the synthesis framework~\cite{pan2025label,ullah2019catheter}. Second, despite recent progress in conditional synthesis, it remains challenging to achieve sufficient diversity in guidewire trajectories and adequate coverage of CAG background complexity, especially under limited annotation budgets. As a result, current augmentation pipelines may not sufficiently cover real CAG examples with complex vessel context and low guidewire visibility. In addition, cost-efficient and scalable generation of guidewire CAG data with directly usable supervision, for example guidewire masks and endpoint coordinates, is still underexplored, limiting the effectiveness of synthetic augmentation for guidewire endpoint localization. Recent CAG-specific studies have further explored geometry-aware data synthesis for stenosis editing and auto-annotation generation for multi-view translational correction, highlighting the growing importance of controllable data generation in coronary angiography~\cite{cao2026automatic,li2026geometrically}.
    
To address these gaps, we propose \textbf{VDSB-GWSyn}, a vessel-constrained diffusion Schr\"{o}dinger bridge framework for controllable guidewire data synthesis on real coronary angiography backgrounds. Our method first learns guidewire shape priors and then synthesizes guidewire masks under vessel segmentation constraints, while directly producing endpoint coordinates. We formulate guidewire synthesis as a targeted inpainting problem so that generation is confined to the masked region and the anatomical background remains undisturbed. We further integrate SPADE to better preserve spatial conditioning information that can otherwise be weakened by normalization layers, improving local alignment between synthesized guidewire appearance and background texture~\cite{park2019semantic}. Finally, we redesign the training objective with region normalization to mitigate gradient dilution caused by extreme guidewire pixel sparsity. Experiments demonstrate that pre-training with VDSB-GWSyn synthetic data greatly improves downstream endpoint localization on real-only evaluation sets.
    
\section{Methods}
    
The proposed VDSB-GWSyn framework is an end-to-end controllable synthesis pipeline designed for coronary angiography (CAG). As illustrated in Fig.~\ref{fig:pipeline}, the framework consists of an Anatomical Mask Generator that samples plausible guidewire geometries and a Mask-Guided Diffusion Schr\"{o}dinger Bridge (DSB) Engine that performs targeted rendering to ensure background consistency.
    
\subsection{Controllable Mask Generation with Anatomical Constraints}
To enable large scale synthesis of anatomically feasible guidewire masks under vessel constraints, we propose a geometry controllable mask generation algorithm.
    
\noindent\textbf{Shape Prior and Radius Modeling.} We learn a global shape prior from clinical masks. Let $\theta \in \mathbb{R}^{2N}$ be a vector representing $N$ skeletonized control points. We extract a low-dimensional manifold $z = \text{PCA}(\theta)$ and model its distribution using a Gaussian Mixture Model (GMM). During sampling, the guidewire radius $r$ is derived from a log-normal prior, $r = \exp(\mathcal{N}(\mu_r, \sigma_r^2))$, and capped by the local vessel diameter to adapt to distal branches.
    
\noindent\textbf{Vessel-Constrained Path Sampling.} Given a vessel mask $\nu$, we construct an 8-connectivity graph $G$ on its skeleton. We utilize Dijkstra's algorithm to search for a base path $P$, where a penalty term is applied to junction nodes (degree $\ge 3$) to suppress unrealistic bifurcations. The algorithm supports precise J-tip control by allowing users to specify a tip location $(x,y)$ and orientation $\phi$. Sampled paths are projected into the vessel lumen using a distance-transform (DT) field. To simulate clinical diversity, we introduce a ``wall-hugging'' parameter $\rho \in [0, 1]$ that controls the target DT value during projection:
\begin{equation}
D_{\mathrm{target}}=(1-\rho)D_{\nu}+\rho d_{\min},
\end{equation}
where $D_{\nu}$ denotes the local vessel DT value and $d_{\min}$ is the minimum clearance to the vessel boundary. Thus, $\rho=0$ keeps the guidewire closer to the vessel centerline, while larger $\rho$ moves it toward the vessel wall without leaving the lumen. This module naturally supports multi-wire synthesis by iteratively sampling disjoint paths within the same anatomical domain.
\subsection{Targeted Inpainting via DSB and SPADE Modulation}
Standard diffusion models tend to redraw the entire image, which can cause texture drift in complex anatomical backgrounds. We therefore formulate guidewire synthesis as a targeted inpainting problem, using the original CAG anatomy as a hard background anchor~\cite{de2021diffusion,lugmayr2022repaint}.
    
\noindent\textbf{Mask-Directed Noise Injection.}
Given a device-free background frame $x_{\mathrm{bg}}$ and a generated guidewire mask $m\in\{0,1\}^{H\times W}$, we construct a noisy conditional endpoint $x_1^{n}$ by confining the perturbation strictly within the mask region:
\begin{equation}
x_1^{n} = (1-m)\odot x_{\mathrm{bg}} + m\odot(x_{\mathrm{bg}} + \sigma\epsilon), \qquad \epsilon\sim\mathcal{N}(0,\mathbf{I}).
\label{eq:mask_noise_injection}
\end{equation}
By keeping the $(1-m)$ background region noise-free, the model treats the surrounding anatomy as a fixed reference. The diffusion Schr\"{o}dinger bridge (DSB)~\cite{de2021diffusion,liu20232} then learns an evolution path that renders guidewire appearance only within the mask $m$.
    
\noindent\textbf{Spatial Feature Alignment.}
To achieve seamless integration between the synthesized guidewire and the original background, we incorporate Spatially-Adaptive Denormalization (SPADE) \cite{park2019semantic}. Specifically, for a normalized feature map $\hat{h}$ at an intermediate layer, SPADE modulates it with spatially varying parameters predicted from the conditioning signal:
\begin{equation}
\mathrm{SPADE}(\hat{h}\,|\,x_1^{n}) = \gamma(x_1^{n})\odot \hat{h} + \beta(x_1^{n}),
\label{eq:spade}
\end{equation}
where $\gamma(\cdot)$ and $\beta(\cdot)$ are learnable functions producing per-location scaling and bias. This modulation encourages the diffusion network to align synthesized guidewire edges with local contrast and noise statistics of the specific CAG frame, while avoiding undesired redrawing of the underlying vascular structure.
    
\subsection{Region-Normalized Objective for Sparse Structures}
Guidewires are extremely sparse in CAG frames, often occupying $<1\%$ of the pixels. To prevent the background from dominating the gradient, we redesign the training objective using region-normalization. The total loss is partitioned into instrument and background components:
\begin{equation}
    \mathcal{L} =
    \lambda_{w}\mathcal{L}_{\mathrm{wire}}+
    \lambda_{b}\mathcal{L}_{\mathrm{bg}},
    \end{equation}
    where
    \begin{equation}
    \mathcal{L}_{\mathrm{wire}} =
    \frac{1}{\sum_i m_i}
    \sum_i m_i \left\|\hat{y}_i-y_i\right\|_2^2 .
    \end{equation}

By setting $\lambda_{w} \gg \lambda_{b}$, we force the model to focus on the high-frequency textures of the guidewire. Furthermore, we introduce a Masked Total Variation (TV) regularization term, acting only on the mask boundary, to enforce structural continuity and mitigate wire fragmentation in low-contrast regions.
    
\begin{figure}[t]
\centering
\includegraphics[width=\textwidth]{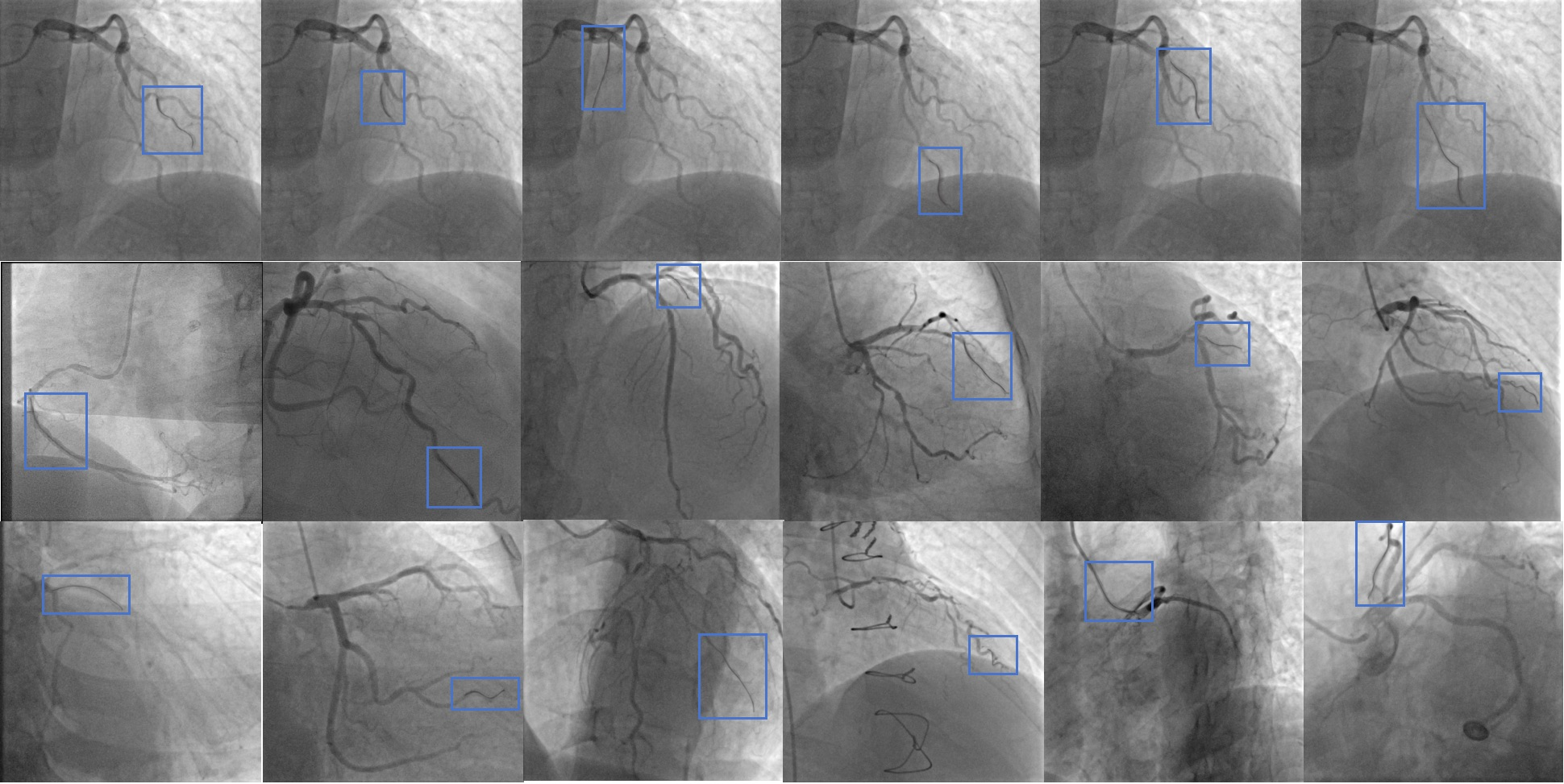}
\caption{\textbf{Qualitative visualization of controllable guidewire synthesis in coronary angiography (CAG).}
\textbf{Top row:} Using the same device-free CAG background, we synthesize guidewires at different vessel locations to demonstrate spatial diversity while preserving the underlying anatomy.
\textbf{Middle row:} Synthesized guidewires under different camera angles, showing robustness to view-dependent appearance variations in CAG.
\textbf{Bottom row:} Synthesized guidewires across backgrounds with different levels of anatomical and noise complexity, illustrating background preservation and realism under challenging fluoroscopic conditions.}
\label{fig:guidewire_qualitative}
\end{figure}
    
\begin{figure}[t]
\centering
\includegraphics[width=0.8\textwidth]{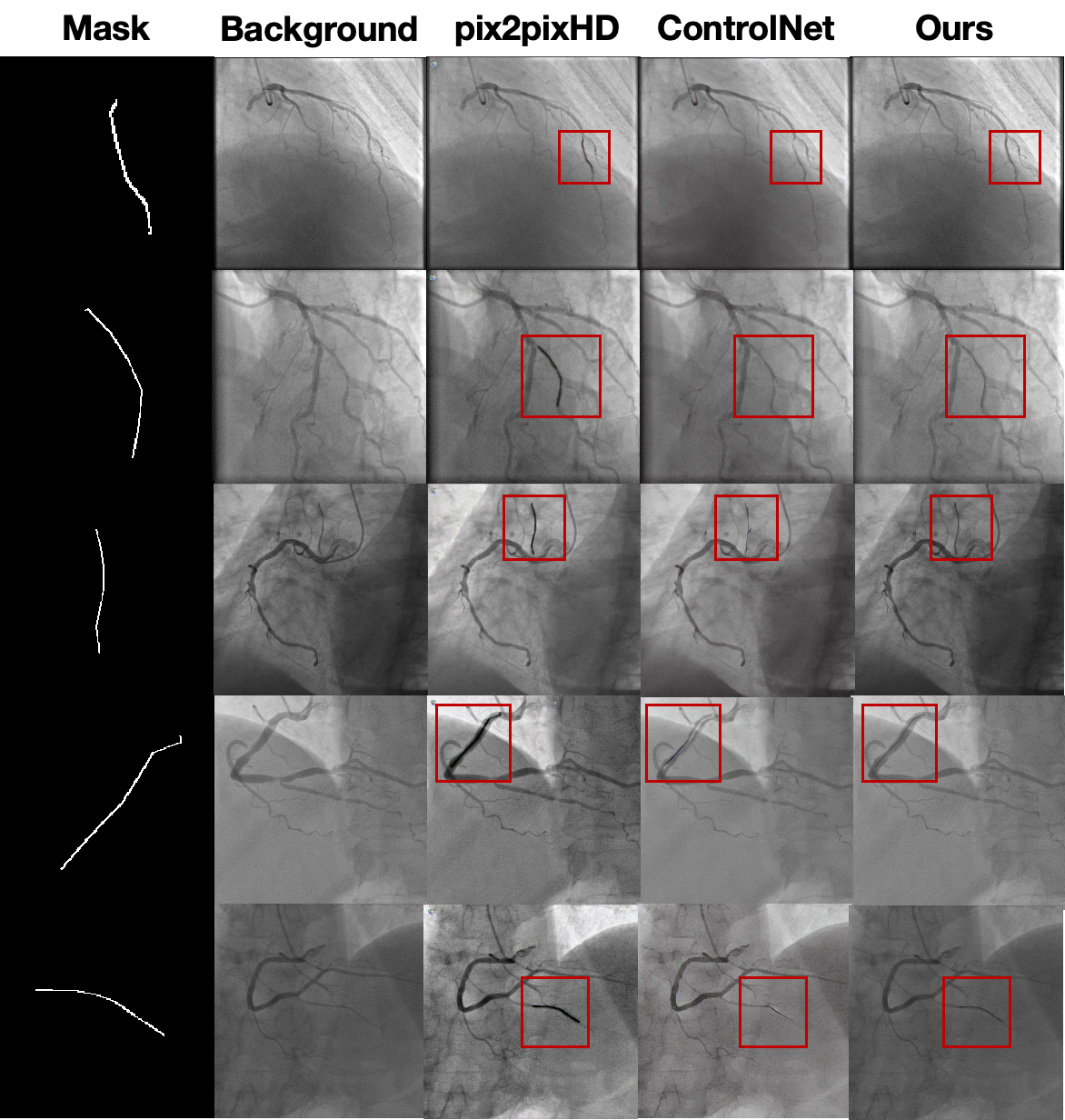}
\caption{\textbf{Qualitative comparison of guidewire synthesis.} Given the same device-free CAG background and the same input guidewire mask, we compare pix2pixHD, ControlNet, and our VDSB-GWSyn. Red boxes indicate the synthesized guidewire regions.}
\label{fig:comparison}
\end{figure} 
      
\section{Experiments}
\subsection{Datasets}
\noindent\textbf{Internal Real Dataset.}
We curate an internal coronary angiography (CAG) dataset consisting of 1{,}125 frames in total. Among them, 283 real frames containing clinical guidewires are manually annotated with pixel-level masks and endpoint coordinates (tip and tail). This annotated subset serves three purposes: (i) extracting guidewire geometry for the shape-prior algorithm, (ii) providing paired samples to train the diffusion renderer, and (iii) acting as the training and evaluation set for the downstream endpoint localization task. For the downstream task, the 283 annotated frames are split into 182/50/51 for training/validation/testing.
    
\noindent\textbf{Vessel Segmentation.}
The CAG images with vessel segmentation masks used for synthesis are drawn from two sources: (i) 955 internal clinical frames whose vessel masks are obtained using U-Mamba~\cite{ma2024u}, and (ii) 832 high-quality frames selected from the ARCADE dataset~\cite{popov2024dataset} (released under the CC0-1.0 license), where vessel masks are generated using the same model and manually filtered for anatomical accuracy. The vessel masks are used exclusively to constrain guidewire mask sampling and to enforce anatomical plausibility; they are not used as inputs to the downstream endpoint localizer.
    
\noindent\textbf{Synthesis Training Set.}
Using 1,787 CAG frames paired with vessel segmentation masks, we first sampled 7,148 candidate guidewire masks (four per frame) and then retained 3,506 for rendering after filtering out anatomically implausible cases, where the sampled trajectories deviated from the vessel course or extended outside the vessel lumen due to segmentation inaccuracies. Then we constructed a 3000-image subset for downstream pre-training via manual quality control, excluding cases where the synthesized guidewire is missing, severely blurred, or strongly fragmented. This screening is applied uniformly and uses only image-level inspection with no downstream model feedback. The synthetic set covers diverse guidewire locations within the same CAG frame, varying clinical camera angles and different levels of background complexity. Representative examples are shown in Fig.~\ref{fig:guidewire_qualitative}. Following the same uniform image-level screening procedure, we also construct a 2,000-image subset for baseline synthetic pre-training to ensure a fair comparison.
    
\subsection{Experimental Setup}
All images are resized to $512 \times 512$. We first fit a guidewire shape prior by skeletonizing clinical masks into 8 control points. The coordinates are compressed via PCA to a 4-dimensional latent space, and the latent distribution is modeled with a 5-component GMM. Our diffusion renderer follows a diffusion Schr\"{o}dinger bridge with 1{,}000 discretization steps, noise levels from $t_{0}=10^{-4}$ to $T=1.0$, and a symmetric beta set to $0.3$. During training, Gaussian noise is injected only inside the guidewire mask regions to enforce localized inpainting. We optimize with AdamW (lr $=5\times 10^{-5}$), batch size $8$, for $20$k iterations, using EMA $0.99$. We additionally apply a masked TV regularizer with weight $0.05$ to preserve thin-structure continuity. For the downstream task, we adopt the RTMPose-S estimator with SimCC label encoding~\cite{jiang2023rtmpose,li2022simcc}. The model is pre-trained on three synthetic sets (1{,}000, 2{,}000, and 3{,}000 images) and then fine-tuned on real data.

\begin{table}[t]
\centering
\caption{Comprehensive comparison with baselines and ablation study on ROI-based generative quality metrics. Lower is better for ROI-FID / ROI-KID; higher is better for IPR (Precision/Recall). Bold indicates the best performance.}
\label{tab:comprehensive_metrics}
\setlength{\tabcolsep}{6pt}
\begin{tabular}{lccc}
\toprule
Method & ROI-KID (\%) $\downarrow$ & IPR (P/R) (\%) $\uparrow$ & ROI-FID $\downarrow$ \\
\midrule
\multicolumn{4}{l}{\textit{Baselines}}\\
pix2pixHD      & $5.5668 \pm 0.1681$ & 31.75 / 45.39 & 72.6365 \\
ControlNet   & $2.1479 \pm 0.1029$ & 35.43 / 50.71 & 53.1129 \\
\midrule
\multicolumn{4}{l}{\textit{Ablation}}\\ 
w/o SPADE \& w/o loss & $2.7071 \pm 0.1693$ & 76.35 / 55.72 & 58.4721 \\
w/o SPADE             & $2.5614 \pm 0.1860$ & 76.93 / 57.01 & 51.6956 \\
\midrule
Ours (DSB + SPADE) & $\mathbf{1.6181 \pm 0.1033}$ & $\mathbf{78.74 / 65.87}$ & $\mathbf{49.1371}$ \\
\bottomrule
\end{tabular}
\end{table}

\begin{table}[t]
\centering
\caption{Downstream guidewire endpoint localization results of our VDSB-GWSyn. Lower is better for MPE; higher is better for PCK.}
\label{tab:downstream_endpoint}
\setlength{\tabcolsep}{6pt}
\begin{tabular}{cccccc}
\toprule
Pre-train data & MPE $\downarrow$ & PCK@3 $\uparrow$ & PCK@6 $\uparrow$ & TIP MPE $\downarrow$ & TAIL MPE $\downarrow$ \\
\midrule
0 & 16.0136 & 52.63 & 78.95 & 19.7241 & 12.3032 \\
1{,}000 & 9.7175 & 77.45 & 84.31 & 9.6383 & 9.7966 \\
2{,}000 & \textbf{7.7126} & \textbf{86.27} & \textbf{91.18} & \textbf{5.8726} & 9.5526 \\
3{,}000 & 7.7781 & 85.29 & 88.24 & 6.7834 & \textbf{8.7727} \\
\bottomrule
\end{tabular}
\end{table}
    
\subsection{Generation Evaluation}
We evaluate the synthesis quality using both distribution alignment and local structure fidelity. Since coronary guidewires are extremely slender, global metrics such as standard FID~\cite{heusel2017gans} and KID~\cite{binkowski2018demystifying} exhibit negligible variation. Following instance-level evaluation~\cite{wang2025grade,xu2025decouple}, we compute FID and KID on mask-centered ROI patches, which are extracted by cropping a tight bounding box around the synthesized guidewire mask, to focus evaluation on guidewire appearance rather than background. We also report IPR~\cite{kynkaanniemi2019improved} to quantify manifold-based precision and recall, reflecting realism and coverage of the real ROI distribution. As shown in Table~\ref{tab:comprehensive_metrics}, our method achieves the best ROI-based performance compared to representative conditional generation baselines, including pix2pixHD~\cite{wang2018high} and ControlNet~\cite{zhang2023adding}. These quantitative trends are consistent with the qualitative comparisons in Fig.~\ref{fig:comparison}. Both baselines are prone to noticeable background drift and inconsistent local contrast, while pix2pixHD often produces overly thick wires with unnatural device-background blending, and ControlNet frequently yields fragmented or vanishing wire segments with weak structural continuity. In contrast, our method preserves background statistics while generating more coherent, anatomically plausible guidewire trajectories with fewer blending artifacts. Ablation results in Table~\ref{tab:comprehensive_metrics} show that ROI-KID improves markedly when incorporating SPADE, indicating that spatially adaptive normalization enhances local feature alignment. Moreover, the proposed region-normalized objective mitigates background-dominated gradients, reducing wire fragmentation and improving structural continuity in synthesized samples.
    
\subsection{Downstream Tasks}
We adopt a synthetic pre-training followed by real fine-tuning protocol to evaluate the effectiveness of synthetic augmentation. RTMPose consistently benefits from the synthetic data. With the largest gain achieved at 2,000 synthesized images, it shows that MPE is reduced from 16.01 px to 7.71 px, and PCK at a 3-pixel threshold increases from 52.63\% to 86.27\%~\cite{yang2012articulated}. For a fair comparison, we apply the same protocol to the baseline methods. We first pre-train RTMPose on 2,000 synthetic images generated by pix2pixHD or ControlNet and then fine-tune the model on the same real training set using identical settings. Both baselines transfer poorly and yield large localization errors, with pix2pixHD reaching an MPE of 34.71 px and PCK@3 px of 5.88\% and ControlNet reaching an MPE of 41.68 px and PCK@3 px of 9.86\%. This gap aligns with their typical synthesis artifacts, where pix2pixHD commonly alters the wire width and local contrast, whereas ControlNet often exhibits instability on slender regions, leading to incomplete wire traces in challenging frames. 
    
\section{Conclusion}
We presented VDSB-GWSyn, a controllable guidewire synthesis framework for coronary angiography, which enables large-scale generation of vessel-constrained guidewire samples under complex coronary backgrounds. The synthesized guidewire samples generated by our model substantially improve endpoint localization on real CAG images. Despite these encouraging results, several limitations remain. In particular, the quality of automatically generated guidewire masks still depends on the accuracy of vessel segmentation. In cases of spatially overlapping vessels or imperfect lumen delineation, sampled trajectories may enter anatomically inconsistent branches, leading to implausible masks that require manual rejection. Our future work will focus on topology-aware constraints and uncertainty-aware sampling to improve robustness and scalability.

\begin{credits}
\subsubsection{\discintname}
The authors have no competing interests to declare that are relevant to the content of this article.
\end{credits}
\bibliographystyle{splncs04}
\bibliography{refs}
\end{document}